\documentclass[conference]{IEEEtran}
\usepackage{amsmath,amssymb,breqn}
\usepackage[linesnumbered,ruled,vlined]{algorithm2e}
\usepackage{xcolor} % Ensure this package is included
\usepackage{multirow} 
\definecolor{darkgreen}{RGB}{50, 150, 100} 

\usepackage{lipsum}
\usepackage[normalem]{ulem}
\useunder{\uline}{\ul}{}
\usepackage{makecell}
\usepackage{color}
\usepackage{graphicx}
\usepackage{subcaption}
\usepackage[    colorlinks=true,
    linkcolor=blue,
    filecolor=blue,      
    urlcolor=blue,
    citecolor=blue,]{hyperref}
\usepackage{hyperref}
\title{\LARGE \bf
Slope Considered Online Nonlinear Trajectory Planning with Differential Energy Model for Autonomous Driving
}

\author{Zhaofeng Tian$^{1}$, Lichen Xia$^{1}$, and Weisong Shi$^{1}$  % <-this % stops a space
\thanks{}% <-this % stops a space
\thanks{$^{1}$The CAR Lab, University of Delaware, Newark, USA
        {\tt\small \{zhaofeng,lxia, weisong\}@udel.edu}}%
}

\begin{document}

\maketitle
\thispagestyle{empty}
\pagestyle{empty}

%%%%%%%%%%%%%%%%%%%%%%%%%%%%%%%%%%%%%%%%%%%%%%%%%%%%%%%%%%%%%%%%%%%%%%%%%%%%%%%%
\begin{abstract}

Achieving energy-efficient trajectory planning for autonomous driving remains a challenge due to the limitations of model-agnostic approaches. This study addresses this gap by introducing an online nonlinear programming trajectory optimization framework that integrates a differentiable energy model into autonomous systems. By leveraging traffic and slope profile predictions within a safety-critical framework, the proposed method enhances fuel efficiency for both sedans and diesel trucks by 3.71\% and 7.15\%, respectively, when compared to traditional model-agnostic quadratic programming techniques. These improvements translate to a potential \$6.14 billion economic benefit for the U.S. trucking industry. This work bridges the gap between model-agnostic autonomous driving and model-aware ECO-driving, highlighting a practical pathway for integrating energy efficiency into real-time trajectory planning.

% The results underscore not only the energy-saving effects but also its feasibility for real-time applications due to its online computability. 

% Both the energy-saving efficacy and online computability of this method highlight the feasibility of real-time applications.

\end{abstract}

\begin{IEEEkeywords}
Trajectory Optimization, Energy Efficiency
\end{IEEEkeywords}
%%%%%%%%%%%%%%%%%%%%%%%%%%%%%%%%%%%%%%%%%%%%%%%%%%%%%%%%%%%%%%%%%%%%%%%%%%%%%%%%

%%%%%%%%%%%%%%%%%%%%%%%%%%%%%%%%%%%%%%%%%%%%%%%%%%%%%%%%%%%%%%%%%%%%%%%%%%%%%%%%

\section{Introduction}

In 2021, truck transportation in the US has covered a remarkable distance of 327.48 billion miles~\cite{truck_mileage}, with fuel costs representing about 30\% of the Total Cost of Ownership (TCO)~\cite{liangkai_fuel, fuel_30}. Despite achieving basic safe operations, current studies and practices in Autonomous Driving (AD) trajectory planning have yet to provide substantial evidence of energy savings under an energy model agnostic framework.  This study proposes a novel \textbf{energy-model-aware} planning paradigm that achieves online nonlinear trajectory optimization on a differentiable energy model.

% According to our study results, leveraging an energy model in autonomous driving could save as much as \$6.14 billion in US trucking fuel expenses, thereby significantly boosting economic efficiency.

In the commonly used AD trajectory generation method, cubic or polynomial curve generation, collision avoidance to both static and dynamic obstacles would be the top priority ~\cite{darpa_motion,state_lattice, poly_trajectory, frenet}, the energy-saving performance is usually considered a byproduct of trajectory smoothing on acceleration or jerk profile~(general energy), instead of an explicit objective under an energy model agnostic frame. 
Although applicable to various vehicle types, using a general energy representation often lacks robust proof for energy optimality. In contrast, Ecological Driving (ECO-driving) specifically assists drivers in operating vehicles within a high-efficiency zone, providing promising insights for energy-efficient autonomous driving that leverages precise energy models~(model-based) for trajectory optimization. To explore the potential of incorporating ECO-driving strategies, related works are investigated below.

Pressing and releasing the gas pedal, vehicles can Pulse and Glide~(PnG), allowing the engine to operate at high-efficiency points~\cite{cao_png}. In converting the PnG problem into a discrete Optimal Control Problem (OCP), variables such as engine rotation speed, torque, and gear ratio are considered. Study~\cite{shengbo_png} solves the discrete OCP using an enumeration strategy and then addresses nonlinear problems. Beyond Internal Combustion Engines~(ICE) vehicles, a Genetic Algorithm (GA) is applied for offline optimization of PnG in Electric Vehicles~(EV), which has proven energy-efficient compared to constant speed driving ~\cite{tian_png}. However, real-time application and safety concerns remain challenges for PnG strategies.

Optimizing a speed profile given a global driving cycle and terrain prior, Eco-driving can achieve better energy efficiency. Some studies consider the improvement produced by taking the terrain info i.e. slope profile into vehicle control~\cite{lookahead_dp1, lookahead_dp2, down_slope, rolling_eco}. Study~\cite{lookahead_dp1} and \cite{lookahead_dp2} use the Dynamic Programming (DP) method to iteratively search an energy-saving speed profiling and gear selection over a lookahead elevation horizon for a heavy diesel truck, and could achieve a fuel consumption reduction of about 3.5\% on the 120 km route without an increase in trip time. In contrast, studies~\cite{down_slope,rolling_eco} avoid optimizing gear selection by using an approximate bivariate polynomial energy consumption function w.r.t. vehicle's velocity and apparent acceleration or power request. Then a slope-considered objective function is solved by establishing the Pontryagin's Maximum principle (PMP) optimal conditions for an OCP. 

Several studies Consider the traffic information, either making the traffic average velocity the desired driving velocity or adjusting ego-car velocity with varying maximum velocity that is imposed by different traffic conditions. Study~\cite{traffic_fuel1} integrates the traffic average speed in the objective function as the desired speed, using an approximated fuel model w.r.t. velocity and acceleration, The PMP method is applied to solve the OCP. Studies~\cite{ocp_scp, osp_scp_itsc} account for the speed limit changes due to varying traffic congestion, and Sequential Quadratic Programming (SQP) is used to optimize the energy consumption over a road slope profile for both an ICE vehicle and an EV. Furthermore, studies~\cite{two_stage1, two_stage2} develop a two-stage method that initially determines a global speed profile and subsequently optimizes a local speed profile adapting to varying traffic flow speed, where Quadratic Programming~(QP) and SQP are used to tackle quadratic and nonlinear problems defined for the local stage. 

Exactly considering intervehicle distance-keeping instead of average traffic flow speed for safety critic requirements, the distance tracking error is formulated in the objective functions of Adaptive Cruise Control~(ACC) system~\cite{acc_dp, acc_pmp1, acc_sqp, down_slope_traffic}. Study~\cite{acc_pmp1} focuses on globally optimizing energy consumption for an electric vehicle (EV) by considering both car-following dynamics and road slope, using Pontryagin's Maximum Principle (PMP) to address the Optimal Control Problem (OCP). Initially, a safe-distance-unconstrained OCP is solved to circumvent the time-dependent boundary value problem introduced by ACC. Subsequently, sections where the distance constraint is violated are identified, and a sub-OCP with a fixed end time is repeatedly solved. Study~\cite{down_slope_traffic} optimize the eco-driving in a Model Predictive Control (MPC) framework, predicting slope profiling and leading car speed over a time horizon, with each timestep incorporating intervehicle distance tracking error into the objective function. Using an approximate continuous ICE fuel model, the OSP over a prediction horizon is solved by computing PMP optimal conditions with a linear solver. Study~\cite{acc_dp} also integrates the ACC distance tracking error as an objective and formulates the DP method to optimize the energy consumption of a Hybrid Electric Vehicle (HEV) over a prediction horizon. Study~\cite{acc_sqp} explores the merits of using SQP to optimize the energy of a HEV in an ACC condition, compared to the high computing costs of PMP, SQP achieves near-optimal results with higher efficiency.

From the experience of the abovementioned studies, we observe that by integrating a model-based energy savings strategy into AD trajectory planning, ECO-driving could provide a rich source of methodologies. However, several practical gaps exist when adapting ECO-driving techniques to real-world autonomous driving applications:

\emph{(i) Hardware Interfaces.} Autonomous vehicles often rely on non-OEM components, meaning the autonomous driving controller may lack direct control over the engine and gearbox. Instead, a lower-level controller executes PID control on the Drive-By-Wire~(DBW) systems for steering, gas, and brakes, as evidenced by DBW industry~\cite{dataspeed} and control commands such as the $AckermannControlCommand$ used in Autoware ~\cite{autoware}. Consequently, plans that involve direct manipulation of gear selection and torque or power requests ~\cite{lookahead_dp1, lookahead_dp2, shengbo_png} become impractical.

\emph{(ii) Onboard Real-time Long Horizon Prediction and Replanning.} Most onboard AD trajectory planning methodologies necessitate frequent iterative predictions and replanning to adapt to rapidly changing conditions and ensure safety. Techniques such as Dynamic Programming (DP) or Pontryagin’s Maximum Principle (PMP) are challenged by long prediction horizons and high replanning frequencies~\cite{ocp_scp, acc_sqp}. Although some studies~\cite{acc_sqp, two_stage1} have utilized Sequential Quadratic Programming (SQP) and Quadratic Programming (QP) to achieve higher computing efficiency, both methods still struggle to meet real-time processing needs, with computation times ranging from 80–496 and 9–28 secondes~\cite{two_stage1}.

\emph{(iii) Safety Critic.} In autonomous driving operations, safety constraints should be strictly adhered to at all times. Studies that only consider average traffic flow speeds or speed limits as references ~\cite{traffic_fuel1, osp_scp_itsc, two_stage2} fail to implement stringent safety measures. Although other research incorporates ACC operations to include distance-keeping errors in the objective function~\cite{acc_dp, acc_pmp1, acc_sqp, down_slope_traffic}, the enforcement of hard safety constraints over the prediction horizon is generally lacking. Methods based on PMP struggle with the computational complexity of enforcing hard constraints, often resorting to incorporating safety measures into the cost function rather than directly constraining state variables~\cite{acc_pmp1}. This approach, while reducing computational demand, potentially compromises the global safety guarantees~\cite{ocp_scp}.

To address the identified gaps in applying ECO-driving techniques to real-world autonomous driving, we present several key contributions:

~\emph{(i) Offline Fuel Model Fitting and Gear Optimization Method.} We propose a method that accurately translates discrete engine characteristics and fuel map data into a bivariate polynomial function w.r.t. velocity and traction acceleration. This approach uses a Non-Linear Programming (NLP) method to quickly solve min-square fitting problems. Unlike previous works~\cite{down_slope, down_slope_traffic} that fit fuel data relative to velocity and apparent acceleration, this study leverages traction acceleration to provide a more accurate fuel rate estimate on rolling roads.

~\emph{(ii) Online NLP-based Trajectory Optimization Method.} We introduce a fast and near-optimal direct NLP optimization method under the ePAC framework that incorporates the energy model and ACC safety constraints into every state along the trajectory, ensuring global safety. This method integrates horizon-flexible traffic-slope predictions into the optimization program, which enables iteratively replanning with real-time frequency on an embedded onboard device.

~\emph{(iii) Comprehensive Testing.} We conduct extensive tests comparing popular online trajectory optimization methods such as Quadratic Programming (QP) and Sequential Quadratic Programming (SQP) across six driving cycles and three different road slope profiles in both a sedan and a truck. Our tests not only measure fuel consumption but also evaluate prediction horizon length, usage of slope prediction, and solving time, which are rarely emphasized previously.

~\emph{(iv) Open-source and Reproducible Implementation.} All our engineering programs are open-sourced (see Appendix), allowing our results to be reproduced without further tuning. This transparency provides a reliable benchmark for the community, fostering further research and development in this area.

These initiatives aim to bridge the current gaps between ECO-driving and energy-model-aware autonomous driving. The rest of the paper is arranged as follows. Offline fuel modeling and gear optimization are introduced in Section~\ref{sec_offline}. The online Nonlinear Trajectory planning method is illustrated in section~\ref{sec_online}. Experiments and Results are shown in Section~\ref{sec_experiments}. The conclusion is drawn in Section~\ref{sec_conclusion}.

\section{Offline Fuel Modeling and Gear Optimization}
\label{sec_offline}
When observing the requirements for AD software platforms like Autoware, control commands sent to the vehicle are usually generalized kinematic signals like steering angle and the linear vehicle speed. For the sake of incorporating an energy model into the AD trajectory planning, the originally discrete engine fuel map w.r.t. torque, engine speed, and gear ratios, will need to be approximated by a function w.r.t. vehicle speed and acceleration, such that even without real-time torque feedback, the upper-level AD controller can achieve a model-based energy optimization.   
\subsection{Basic Dynamics}
In this study, both a sedan and a truck are used for experiments and analysis, where we refer to study~\cite{down_slope} for the sedan parameter settings, and we use a diesel truck's data to illustrate the modeling process since study~\cite{down_slope} does not provide the original engine and fuel map data. Some necessary vehicle dynamic and fuel consumption formulas are introduced below for later fuel rate modeling.
\begin{figure}
    \centering
    \begin{subfigure}{.55\textwidth}
        \centering
        \includegraphics[width=.8\linewidth]{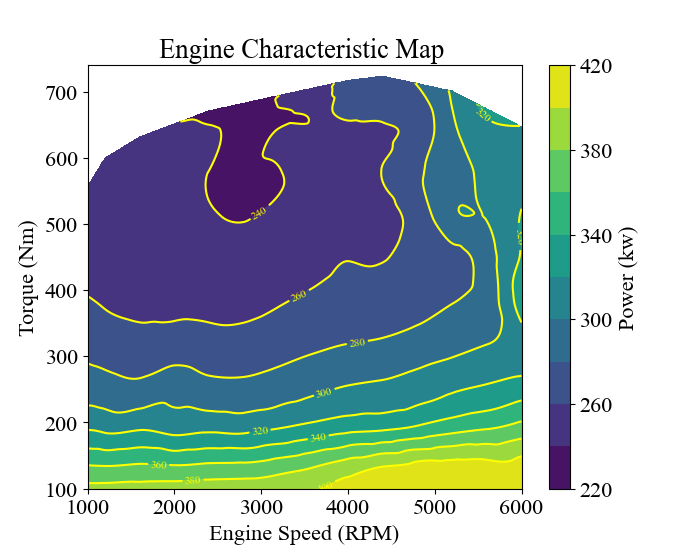}
        \caption{Engine characteristic map}
        \label{engine}
    \end{subfigure}
    
    \begin{subfigure}{.55\textwidth}
        \centering
        \includegraphics[width=.8\linewidth]{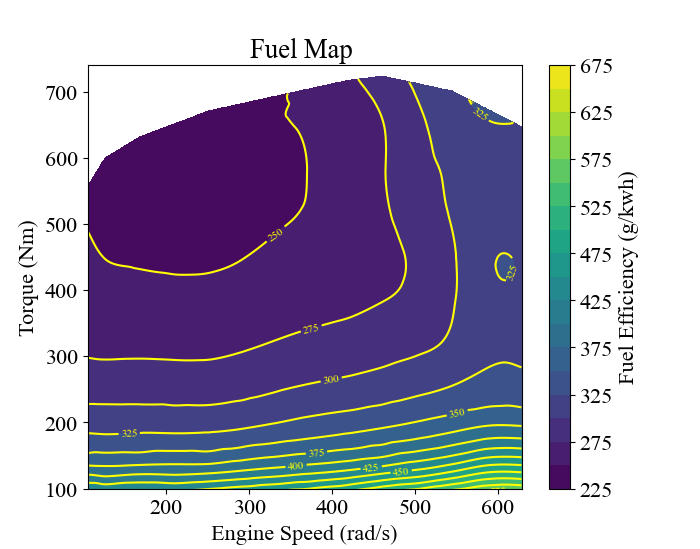}
        \caption{Fuel efficiency map}
        \label{fuel_map}
    \end{subfigure}

    \caption{Engine characteristic and fuel efficiency maps of the diesel truck used in this study.}
\end{figure}

\begin{figure}
    \centering
    
    \begin{subfigure}{.55\textwidth}
        \centering
        \includegraphics[width=.8\linewidth]{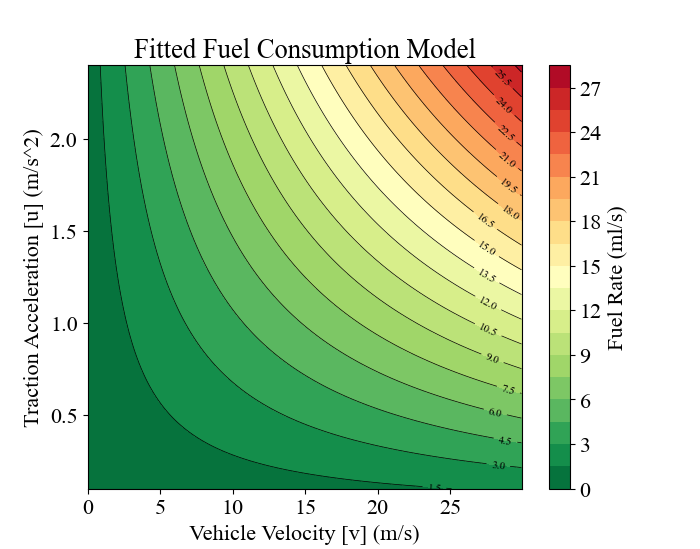}
        \caption{Fitted fuel model}
        \label{fitted}
    \end{subfigure}
    
    \begin{subfigure}{.55\textwidth}
        \centering
        \includegraphics[width=.8\linewidth]{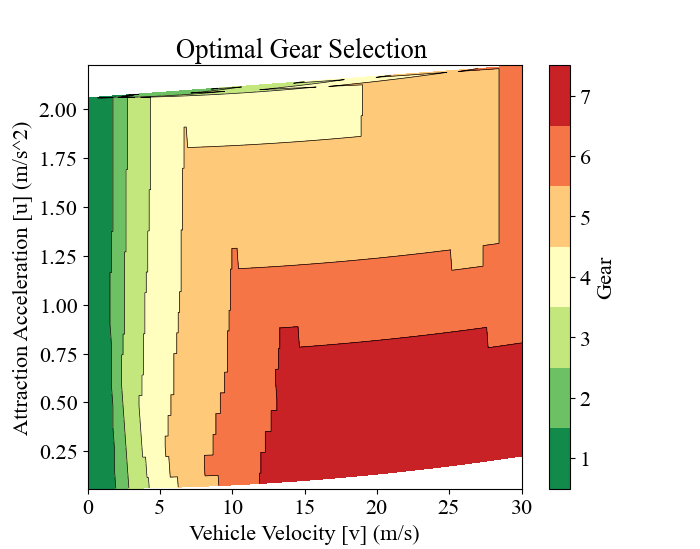}
        \caption{Optimized gear selection}
        \label{gear}
    \end{subfigure}
    
    \caption{Fitted fuel model and optimized gear selection with respect to $u$ and $v$ for the 7-speed truck.}
\end{figure}

\begin{equation}
    k_1 = \frac{C_d \cdot \rho \cdot A_v}{2 \cdot M}
\end{equation}

\begin{equation}
    k_2 = \mu \cdot g , \quad k_3 = g
\end{equation}

\begin{equation}
    \label{ar}
    a_R = k_1 \cdot v^2 + k_2 \cdot cos(\theta(s)) + k_3 \cdot sin(\theta(s))
\end{equation}

\begin{equation}
\label{eq_u}
    \underbrace{u = a_V +  a_R + a_B}_{\text{vehicle dynamic equation}}
\end{equation}

\begin{equation}
    \label{eq_sv}
     \underbrace{v = v_0 + \int_{t_0}^{t_{k}}a \, \Delta t, \quad s = s_0 + \int_{t_0}^{t_{k}}v \, \Delta t}_{\text{vehicle kinematic equations}} 
\end{equation}

\begin{equation}
\label{eq_ft}
    F_t = u \cdot M, \quad T_w = F_t \cdot r
\end{equation}
\begin{equation}
    \label{eq_te}
      T_e = \frac{T_w }{i_t \cdot \eta},  \quad \omega_e = \frac{v \cdot i_t }{r}, \quad i_t = i_g \cdot i_f  
\end{equation}
\begin{equation}
\label{eq_pe}
    P_e = \mathcal{F}_P(\omega_e, T_e), \quad \eta_{f} = \mathcal{F}_{fe}(\omega_e, T_e)   
\end{equation}
    
\begin{equation}
\label{eq_fr}
    f_r = \frac{P_e \cdot \eta_f}{c_u} 
\end{equation}
Where $C_d$, $\rho$, $A_v$, and $M$ respectively denote air drag coefficient, air density frontal area, and equivalent mass; $\mu$ and $g$ are the rolling resistance coefficient and gravity value; $s$, $v$, $u$, are the vehicle's path coordinate, velocity, and traction acceleration; $a_V$, $a_R$, $a_B$ denote apparent; acceleration, resistance acceleration, and brake acceleration; Road slope/grade is denoted by $\theta(s)$, which generally can be obtained from road profiling or GPS information from road coordinate $s$~\cite{down_slope, rolling_eco}. 

To acquire the fuel rate, referring to Equation~(\ref{eq_ft}-\ref{eq_fr}), engine torque $T_e$ can be derived from wheel torque $T_w$ with traction force $F_t$,  wheel radius $r$, the transmission ratio $i_t$, and efficiency $\eta_t$, in which $i_t$ equals to the product of gear ratio $i_g$ and final drive ratio $i_f$, where traction force $F_t$ is directly related to $u$; Engine speed $\omega_e$ can be calculated with vehicle speed $v$ and transmission ratio $i_t$; By using two interpolated functions $\mathcal{F}_P$ and $\mathcal{F}_{fe}$, engine power $P_e$(w) and fuel efficiency $\eta_f$ (g/kwh) can be quired form $\omega_e$, $T_e$, as shown in Fig. \ref{engine} and \ref{fuel_map}; A fuel rate $fr$ at unit (ml/s) referring to study~\cite{down_slope}, can be derived from the product of $P_e$ and $\eta_f$ with a coefficient $c_u = \rho_g \cdot 1000 \cdot 3600$ that converts the consumption rate unit to (ml/s), where $\rho_g = 0.85$ (g/ml) represents the disel density.

\subsection{Fuel Rate Modeling}
Previous works tend to approximate the fuel consumption map into a bivariate quadralic~\cite{va_fuel_model} or polynomial~\cite{down_slope, down_slope_traffic} functions with vehicle speed $v$ and apparent acceleration $a$ as two variables, achieving a differentiable numerical optimization. Considering the traction acceleration $u$ would involve fuel consumption in a more direct way than apparent acceleration $a$ as slope $\theta(\cdot)$ is also incorporated referring to equation~(\ref{eq_u}), we fit the fuel model and optimize the gear w.r.t. $v$ and $u$. With a similar fitting function~\cite{down_slope} with $a_V$ replaced by $u$ in Equation~(\ref{eq_fit}).
\begin{equation}
    \label{eq_fit}
     \hat{f_r}(v, u) = o_0 + o_1v +o_2v^2 + o_3v^3 + o_4v^4 + (c_0+c_1v+c_2v^2)u
\end{equation}

 To precisely fit a fuel rate function, the gear selection policy needs to be optimized in advance to eliminate the impact introduced by different engine working conditions caused by different gear selections, such that the optimized gear selection policy will let the engine work on high-efficiency points. The pseudocode of our implementation is described in Algorithm~\ref{ag_gear}, where the returned $data\_list$ contains $v$, $u$ sampling data with the determined optimal fuel rate and gear number $opt\_f_r$ and $opt_gear$. $\omega_e, T_e$ are bounded by $[62 ~\text{(idling)}, 630]$ rad/s and $[0, 724]$ Nm.

With the optimized gear selection as Fig.~\ref{gear} shows and $v$, $u$ samples in $data\_list$, we now can design a min-square objective function that fits the parameters in Equation~(\ref{eq_fit}) as below to minimize the squared error over all samples:
\begin{equation}
J_{fit} = \min_{o_0, o_1, o_2, o_3, o_4, c_0, c_1, c_2} \sum_{i=1}^{N} \left( f_r^{(i)} - \hat{f_r}^{(i)} \right)^2
\end{equation}

where $f_r^{(i)}$ is the observed fuel rate in the \(i\)-th data tuple of $data\_list$, \(\hat{f_r}^{(i)}\) is the estimated fuel rate using the model for the \(i\)-th data point, and \(N\) is the total number of data points. The min-square problem can be efficiently solved using the nonlinear optimization method IPOPT~\cite{ipopt} within a Casadi~\cite{casadi} symbolic problem-setting interface. The fitted fuel estimated model can limit the average prediction error at $0.1047$ (ml/s) on $data\_list$, especially for this truck, within the more general working conditions $v \sim (5, 25)$ $u \sim (0.1, 1.0)$, the error will be controlled within $0.0596$ (ml/s). The fuel rate w.r.t. $u$ and $v$ is mapped as Fig.~\ref{fitted}. The optimized parameters $o_1$-$o_4$, $c_0$-$c_2$ are recorded in Appendix.

\begin{algorithm}
\caption{Optimize Gear Selection }
\label{ag_gear}
\SetAlgoLined
% \KwData{$v\_range$, $a\_range$, $gear\_list$, constants $k1, k2, k3, M, wheel\_r, final\_drive, trans\_eff$}
% \KwResult{Returns the optimized gear selection data in $data\_list$}
Initialize $data\_list$ $\leftarrow$ $\emptyset$\;
\For{each $v$ in $v\_range$}{
    \For{each $a$ in $a\_range$}{
        % $u \leftarrow a + k1 \cdot v^2 + k2 $;
        % $F \leftarrow u \cdot M$ and $Tw \leftarrow F \cdot wheel\_r$\;
        % $opt\_f_r \leftarrow \infty$, $opt\_gear \leftarrow 0$\;
        Compute $u$, $F_t$, $T_w$ ~(\ref{eq_u})~(\ref{eq_ft})\;
        \For{each $gear$ in $gear\_list$}{
            Calculate $T_e$, $\omega_e$, with gear ratio~(\ref{eq_te}) \;
            Calculate $P_e$ and $f_r$~(\ref{eq_pe})~(\ref{eq_fr})\;
            \If{$f_r < opt\_fr$}{
                $opt\_f_r \leftarrow f_r$, $opt\_gear \leftarrow gear$\;
            }
        }
        \If{$opt\_gear \neq 0$}{
            Append $(v,u,opt\_f_r,opt\_gear)$ to  $data\_list$\;
        }
    }
}
\Return{$data\_list$}\;

\end{algorithm}

\section{Nonlinear Online Trajectory Optimization}
\label{sec_online}
In this section, we present an online trajectory optimization framework that considers the prediction of the leading vehicle's trajectory and the road slope profile over the prediction time horizon. Also, the proposed framework allows both convex/linear-quadratic and nonconvex/nonlinear problem setups without extra effort. For the convex setting, we set up a QP problem for the general-energy based trajectory planning method that is widely used in AD studies~\cite{frenet, poly_trajectory}; For the nonconvex setting, we directly optimize the objective on the fuel rate model, and subsequently, an SQP and NLP settings are adapted to tackle the challenges posed by the highly-nonlinear fuel rate estimate model defined in Equation~(\ref{eq_fr}).

\subsection{Traffic and Road Profile}
In this study, we focus on trajectory planning in the longitudinal dimension, where the traffic information is considered as the leading car profile, such the ego agent should plan a safe trajectory to keep an appropriate distance from the leading car to avoid potential collisions. This study define the leading car profile as a driving cycle~\cite{acc_pmp1,acc_dp},  including travel distance profile, speed profile, and acceleration profile. The leading agent and eco-agent states can be modeled as follows.
\begin{equation}
    \label{eq_states}
    \underbrace{x_l = [s_l, v_l, a_{Vl}]}_{\text{leading agent states}}, \quad \underbrace{x = [s, v, u, a_V, a_B]}_{\text{ego agent states}}
\end{equation}

\begin{equation}
\label{eq_trajectory}
  \mathbf{X} = \left\{ x^{t_i} = \begin{bmatrix} \mathbf{S} = \{s^{t_i}\} \\ \mathbf{V}=\{v^{t_i}\} \\ \mathbf{U}=\{u^{t_i}\} \\ \mathbf{A}=\{a_V^{t_i}\} \\ \mathbf{B}=\{a_B^{t_i}\} \end{bmatrix} \bigg| i \in [0, N_T], t_{i+1} - t_i = \Delta t \right\}  
\end{equation}

Where the leading agent state is defined as a 3-tuple, and the ego agent consists of 5-tuple states since dynamic states like traction and brake acceleration $u$, and $a_B$ are also included~(\ref{eq_states}). Given a prediction horizon $H$ with a second unit, the discretized timestep total number $N_T = \text{int}(\frac{H}{\Delta t})$ with a minimum time granularity $\Delta t$, the predicted trajectory $\mathbf{X}$ of the ego car consists of the $\mathbf{S, V, U, A, B}$ trajectories of their respective variable over the prediction horizon as Equation~(\ref{eq_trajectory}) shows. Similarly, the predicted leading car trajectory $\mathbf{X_l}$ can be formulated with its state trajectories $\mathbf{S_l, V_l, A_l}$ correspondingly. The intervehicle safety ACC constraint can be formulated by the below.
\begin{equation}
    \label{eq_acc}
      d_{max} > \mathbf{S_l} - (\mathbf{S} + t_h \cdot \mathbf{V}) > d_{min} \quad \text{(ACC constraints)}
\end{equation}
Where $t_h$ is a time headway, $d_{min}$ is the minimum distance that two agents should keep at zero-speed conditions, whereas $d_{max}$ is also imposed to constrain the travel distance of two cars within a proximate range, and facilitate a consistent basis for the comparison of different methods. For road profiling, we leverage a mixed trigonometric equation to formulate both slope and elevation profiles shown below.
\begin{equation}
\label{eq_theta}
 \theta(s) = \theta_0 + \sum_{i=1}^{N}a_{gi}\cdot sin(\frac{2\pi s}{ l_{wi}})   
\end{equation}
\begin{equation}
    \label{eq_elevation}
    h = h_0 + \int_{s_s}^{s_e} \mathrm{tan} \, \theta(s) \, ds
\end{equation}
Road slope profiles are formulated along road longitudinal coordinate $s$ by a static slope constant to simulate constant up or down hills, and a $N$-component mixture of $sin(\cdot)$ waves that $i$-th wave is parameterized by an amplitude $a_{gi}$ and a wavelength $l_{wi}$; The elevation is profiled with an initial constant $h_0$ and integration of $tan\,\theta(s)$ along the $s$.

\subsection{Optimization Problem Setup}
As mentioned above, we set up the trajectory planning problem as an optimization problem in three ways, i.e., QP, SQP, and NLP, which will be illustrated subsequently. Some commonly used kinematic constraints are formulated first.
\begin{equation}
\label{eq_kinematic}
\left\{
\begin{aligned} 
    \mathbf{S}[i+1] &= \mathbf{S}[i] +\mathbf{V}[i] \,\Delta t + \frac{1}{2}\,\mathbf{A}[i]\,\Delta t^2\,\\ 
    \mathbf{V}[i+1] &= \mathbf{V}[i] + \mathbf{A}[i]\,\Delta t \quad \text{(kinematic constraints)}\\
\end{aligned}
\right.
\end{equation}

Dynamic constraints should also be considered for SQP and NLP settings, which can be formulated as below.
\begin{equation}
\label{eq_dynamic}
\left\{
\begin{aligned} 
    \mathbf{R} &= k_1 \cdot \mathbf{V}^2 + k_2 \cdot \text{cos}(\mathbf{G}) + k_3 \cdot \text{sin}(\mathbf{G}) \\ 
    \mathbf{U} &= \mathbf{A} + \mathbf{R} + \mathbf{B}  \quad \text{(dynamic constraints)}\\
\end{aligned}
\right.
\end{equation}
Where $\mathbf{R}$ is a trajectory array of resistance acceleration $a_R$, and $\textbf{G}$ is a predicted road slope profile shown below.
\begin{equation}
    \label{eq_g}
    \textbf{G} = \{ \theta(s^{t_i})\,|\, i \in [0, N_T] \}
\end{equation}

\subsubsection{QP}
A QP problem is set up for the general-energy based objectives that are popularly used in autonomous driving trajectory planning.
\begin{equation}
\begin{aligned}
    \label{eq_qp}
    \min_{\mathbf{A}} \quad & J = w_1 \cdot \| \mathbf{V_l} - \mathbf{V} \|_2^2 \, + \, w_2 \cdot \| \mathbf{A} \|_2^2 \\
    \text{s.t.} \quad & \text{Constraints~(\ref{eq_acc})~(\ref{eq_kinematic})} \\
    & \text{and} \quad \mathbf{V, A} \quad \text{bounds} \\
\end{aligned}
\end{equation}
Where $w_1, w_2$ are the weights of two objectives, and the first item is the velocity reference objective expecting the ego agent could keep a similar velocity with the leading agent. Making the reference speed the maximum speed could force the value of the first item to be comparatively too large when the leading is running in a low-speed zone, such that the ACC safety would solely rely on the hard constraints, and the fuel economy would be harmed. The second item is the general energy objective aimed at minimizing the acceleration costs to improve both smoothness and fuel economy. 

\subsubsection{SQP and NLP}
SQP converts a nonlinear/nonconvex problem into iterative convex problems that can be solved by QP, while a direct NLP method can handle the nonlinear/nonconvex objectives and dynamics without convex approximation like SQP. The original nonlinear problem setting tackled by a direct NLP method is expressed below.
\begin{equation}
    \label{eq_fuel_obj}
    \begin{aligned}     
    \mathbf{F(V, U)} & = o_0 + o_1\mathbf{V} +o_2\mathbf{V}^2 + o_3\mathbf{V}^3 + o_4\mathbf{V}^4 \\ &+ (c_0+c_1\mathbf{V}+c_2\mathbf{V}^2)\cdot \mathbf{U} \\
    & = \{\hat{f_r}^{t_i}(\cdot)\,|\,i \in [0, N_T]\}\\
    \end{aligned}
\end{equation}

\begin{equation}
\begin{aligned}
    \label{eq_nlp}
    \min_{\mathbf{U,B}} \quad & J = w_1 \cdot \| \mathbf{V_l} - \mathbf{V} \|_2^2 \, + \, w_2 \cdot \| \mathbf{A} \|_2^2 \\ &+ w_2 \cdot \| \mathbf{B} \|_2^2 \, + \, w_3 \cdot \|\mathbf{F(\cdot)} \|_1 \\  
    \text{s.t.} \quad & \text{Constraints~(\ref{eq_acc})~(\ref{eq_kinematic})~(\ref{eq_dynamic})} \\
    & \text{and} \quad \mathbf{V, A, U, B} \quad \text{bounds} \\
\end{aligned}
\end{equation}

Where the 1-norm of fuel rate trajectory $\mathbf{F(\cdot)}$ is integrated into the objective function referring to Equation~(\ref{eq_fuel_obj})(\ref{eq_nlp}) as a model-aware paradigm, and a new control variable $\mathbf{B}$ is introduced in objectives to comply with vehicle dynamics. To leverage the SQP method, the original NLP problem setting needs to be approximated to a QP setting by Taylor Expansion such that the QP solver can iteratively solve the approximated QP problems. Where the nonlinear fuel rate trajectory array $\mathbf{F}$ needs to be approximate by $\mathbf{\Tilde{F}}$, a quadratic or linear form based on the Taylor Expansion on the reference trajectory $\mathbf{\bar{V}, \bar{U}}$. Note that in the expression below, we omit the subscripts in partial derivatives e.g.,$\frac{\partial{\mathbf{F}}}{\partial{\mathbf{V}}\partial{\mathbf{U}}} |_{\bar{\mathbf{V}}, \bar{\mathbf{U}}}$.

\begin{equation}
\label{eq_approx}
    \begin{aligned}
     \mathbf{\Tilde{F}}(\mathbf{V,U}) & =\mathbf{F}(\mathbf{\bar V, \bar U})+\frac{\partial \mathbf{F}}{\mathbf{\partial V}}(\mathbf{V- \bar V}) +\frac{\partial \mathbf{F}}{\mathbf{\partial U}}(\mathbf{U-\bar U})  \\
     & +\frac{\partial^2\mathbf{F}}{\mathbf{\partial V \partial U}}(\mathbf{V - \bar V}) (\mathbf{U - \bar U}) \\ 
     &+ \frac{1}{2}\frac{\partial^2\mathbf{F}}{\mathbf{\partial U}^2}(\mathbf{U-\bar U})^2 + \frac{1}{2}\frac{\partial^2\mathbf{F}}{\partial \mathbf{V}^2}(\mathbf{V-\bar V})^2 \\
\end{aligned}
\end{equation}
Moreover, on the constraint side, the QP setting requires a linear state transition, consequently $\mathbf{R}$ needs to be replaced by an approximation on reference $\mathbf{\bar V}$ that $\mathbf{\Tilde{R}} = k_1(\mathbf{\bar{V}}^2 + 2\mathbf{\bar V}(\mathbf{V - \bar V})) + (...)$. Then given an initial reference $\mathbf{\bar V, \bar U}$ guess, SQP gets a warm start and iteratively updates the reference with the solution solved last iteration, such the whole optimization process resembles a Newton method. Note although jerk is not included in the optimization for generalizability when updating the agent's states, jerk is hard constrained within $1 m/s^3 $.

\section{Experiments}
\label{sec_experiments}

\begin{table*}[]
\caption{Comprehensive Tests}
\centering
\begin{tabular}{cccccccccc}
\hline
\makecell[c]{\textbf{Agent} \\ \textbf{ID}} & \makecell[c]{\textbf{Travel} \\ \textbf{Time} \\ \textbf{(s)}} & \makecell[c]{\textbf{Travel} \\ \textbf{Distance} \\ \textbf{(m)}} & \makecell[c]{\textbf{Fuel} \\ \textbf{Consumption} \\ \textbf{(ml)}} & \makecell[c]{\textbf{Average} \\ \textbf{Fuel Rate} \\ \textbf{(ml/s)}} & \makecell[c]{\textbf{Average} \\ \textbf{Fuel} \\ \textbf{Efficiency} \\ \textbf{(L/100km)}} & \makecell[c]{\textbf{Average} \\ \textbf{Speed} \\ \textbf{(m/s)}} & \makecell[c]{\textbf{Fuel Efficiency} \\ \textbf{Improvement} \\ \textbf{(\%)} } & \makecell[c]{\textbf{Average} \\ \textbf{Solving} \\ \textbf{Time A } \\ \textbf{(ms)}}  & \makecell[c]{\textbf{Average} \\ \textbf{Solving} \\ \textbf{Time B } \\ \textbf{(ms)}} \\ \hline
%%%%%%%%%%%%%%%%%%%%%%%%%%%%%%%%
\makecell[c]{Truck-L} & \makecell[c]{13477.20} & \makecell[c]{123177.83} & \makecell[c]{20525.89} & \makecell[c]{1.5230} & \makecell[c]{16.6636} & \makecell[c]{9.1397} & \makecell[c]{0.00} & \makecell[c]{-} & \makecell[c]{-} \\ \hline
\makecell[c]{Truck-QP-5} & \makecell[c]{13477.20} & \makecell[c]{123058.27} & \makecell[c]{18861.42} & \makecell[c]{1.3995} & \makecell[c]{15.3272} & \makecell[c]{\textbf{9.1308}} & \makecell[c]{8.02} & \makecell[c]{ 1.1828} & \makecell[c]{4.4574} \\
\makecell[c]{Truck-SQP-5} & \makecell[c]{13477.20} & \makecell[c]{121924.53} & \makecell[c]{19398.65} & \makecell[c]{1.4394} & \makecell[c]{15.9104} & \makecell[c]{9.0467} & \makecell[c]{4.52} & \makecell[c]{5.3067} & \makecell[c]{20.0371} \\
\makecell[c]{Truck-NLP-5} & \makecell[c]{13477.20} & \makecell[c]{119981.86} & \makecell[c]{17074.13} & \makecell[c]{1.2669} & \makecell[c]{14.2306} & \makecell[c]{8.9026} & \makecell[c]{\color{darkgreen}{\textbf{14.60}}} & \makecell[c]{10.9702} & \makecell[c]{40.7752} \\ \hline
%%%%%%%%%%%%%%%%%%%%%%%%%%%%%%%%%%%%
\makecell[c]{Truck-QP-10} & \makecell[c]{13477.20} & \makecell[c]{122871.33} & \makecell[c]{19497.51} & \makecell[c]{1.4467} & \makecell[c]{15.8682} & \makecell[c]{9.1170} & \makecell[c]{4.77} & \makecell[c]{1.6621} & \makecell[c]{-} \\

\makecell[c]{\textcolor{gray}{Truck-SQP-10}} & \makecell[c]{\textcolor{gray}{10942.50}} & \makecell[c]{\textcolor{gray}{104411.57}} & \makecell[c]{\textcolor{gray}{16726.72}} & \makecell[c]{\textcolor{gray}{1.5286}} & \makecell[c]{\textcolor{gray}{16.0199}} & \makecell[c]{\textcolor{gray}{9.54}} & \makecell[c]{\textcolor{gray}{3.86}} & \makecell[c]{\textcolor{gray}{8.6851}} & \makecell[c]{\textcolor{gray}{-}} \\

\makecell[c]{Truck-NLP-10} & \makecell[c]{13477.20} & \makecell[c]{119677.71} & \makecell[c]{17439.35} & \makecell[c]{1.2940} & \makecell[c]{14.5719} & \makecell[c]{8.8800} & \makecell[c]{\textbf{12.55}} & \makecell[c]{18.5562} & \makecell[c]{-} \\ \hline
%%%%%%%%%%%%%%%%%%%%%%%%
\makecell[c]{Truck-QP-5F} & \makecell[c]{13477.20} & \makecell[c]{123058.27} & \makecell[c]{18861.42} & \makecell[c]{1.3995} & \makecell[c]{15.3272} & \makecell[c]{\textbf{9.1308}} & \makecell[c]{8.02} & \makecell[c]{1.1263} & \makecell[c]{-} \\

\makecell[c]{\textcolor{gray}{Truck-SQP-5F}} & \makecell[c]{\textcolor{gray}{4913.30}} & \makecell[c]{\textcolor{gray}{75300.29}} & \makecell[c]{\textcolor{gray}{11227.98}} & \makecell[c]{\textcolor{gray}{2.2852}} & \makecell[c]{\textcolor{gray}{14.9109}} & \makecell[c]{\textcolor{gray}{15.3258}} & \makecell[c]{\textcolor{gray}{10.52}} & \makecell[c]{\textcolor{gray}{4.7907}} & \makecell[c]{\textcolor{gray}{-}} \\ 

\makecell[c]{Truck-NLP-5F} & \makecell[c]{13477.20} & \makecell[c]{119987.92} & \makecell[c]{17079.31} & \makecell[c]{1.2673} & \makecell[c]{14.2342} & \makecell[c]{8.9030} & \makecell[c]{\textbf{14.58}} & \makecell[c]{10.8604} & \makecell[c]{-} \\ \hline \hline
%%%%%%%%%%%%%%%%%%%%%%%%%%%%%
\makecell[c]{Sedan-L} & \makecell[c]{13477.20} & \makecell[c]{123177.83} & \makecell[c]{8329.70} & \makecell[c]{0.6181} & \makecell[c]{6.7623} & \makecell[c]{9.1397} & \makecell[c]{0.00} & \makecell[c]{-} & \makecell[c]{-} \\ \hline
\makecell[c]{Sedan-QP-5} & \makecell[c]{13477.20} & \makecell[c]{122265.88} & \makecell[c]{7814.52} & \makecell[c]{0.5798} & \makecell[c]{6.3914} & \makecell[c]{9.0721} & \makecell[c]{5.49} & \makecell[c]{1.1952} & \makecell[c]{4.4082} \\
\makecell[c]{Sedan-SQP-5} & \makecell[c]{13477.20} & \makecell[c]{120890.79} & \makecell[c]{7867.88} & \makecell[c]{0.5838} & \makecell[c]{6.5083} & \makecell[c]{8.9700} & \makecell[c]{3.76} & \makecell[c]{5.3042} & \makecell[c]{20.1379} \\
\makecell[c]{Sedan-NLP-5} & \makecell[c]{13477.20} & \makecell[c]{119719.42} & \makecell[c]{7368.26} & \makecell[c]{0.5467} & \makecell[c]{6.1546} & \makecell[c]{8.8831} & \makecell[c]{\color{darkgreen}{\textbf{8.99}}} & \makecell[c]{10.4392} & \makecell[c]{39.2654} \\ \hline
%%%%%%%%%%%%%%%%%%%%%%%
\makecell[c]{Sedan-QP-10} & \makecell[c]{13477.20} & \makecell[c]{122912.78} & \makecell[c]{7996.89} & \makecell[c]{0.5934} & \makecell[c]{6.5061} & \makecell[c]{\textbf{9.1201}} & \makecell[c]{3.79} & \makecell[c]{1.6399} & \makecell[c]{-} \\
\makecell[c]{Sedan-SQP-10} & \makecell[c]{13477.20} & \makecell[c]{120802.82} & \makecell[c]{7878.56} & \makecell[c]{0.5846} & \makecell[c]{6.5218} & \makecell[c]{8.9635} & \makecell[c]{3.56} & \makecell[c]{9.0283} & \makecell[c]{-} \\
\makecell[c]{Sedan-NLP-10} & \makecell[c]{13477.20} & \makecell[c]{119437.12} & \makecell[c]{7432.23} & \makecell[c]{0.5515} & \makecell[c]{6.2227} & \makecell[c]{8.8622} & \makecell[c]{\textbf{7.98}} & \makecell[c]{18.5075} & \makecell[c]{-} \\ \hline
%%%%%%%%%%%%%%%%%%%%%
\makecell[c]{Sedan-QP-5F} & \makecell[c]{13477.20} & \makecell[c]{122265.88} & \makecell[c]{7814.52} & \makecell[c]{0.5798} & \makecell[c]{6.3914} & \makecell[c]{9.0721} & \makecell[c]{5.49} & \makecell[c]{1.1279} & \makecell[c]{-} \\
\makecell[c]{\textcolor{gray}{Sedan-SQP-5F}} & \makecell[c]{\textcolor{gray}{2213.70}} & \makecell[c]{\textcolor{gray}{48671.63}} & \makecell[c]{\textcolor{gray}{2935.05}} & \makecell[c]{\textcolor{gray}{1.3259}} & \makecell[c]{\textcolor{gray}{6.0303}} & \makecell[c]{\textcolor{gray}{21.99}} & \makecell[c]{\textcolor{gray}{10.82}} & \makecell[c]{\textcolor{gray}{6.1508}} & \makecell[c]{\textcolor{gray}{-}} \\ 

\makecell[c]{Sedan-NLP-5F} & \makecell[c]{13477.20} & \makecell[c]{119723.33} & \makecell[c]{7368.78} & \makecell[c]{0.5468} & \makecell[c]{6.1548} & \makecell[c]{8.8834} & \makecell[c]{\textbf{8.98}} & \makecell[c]{10.5095} & \makecell[c]{-} \\ \hline

\end{tabular}
\label{tab_experiments}
\end{table*}

\begin{figure}[h]

    \centering
    \begin{subfigure}{.50\textwidth}
        \centering
        \includegraphics[width=\linewidth]{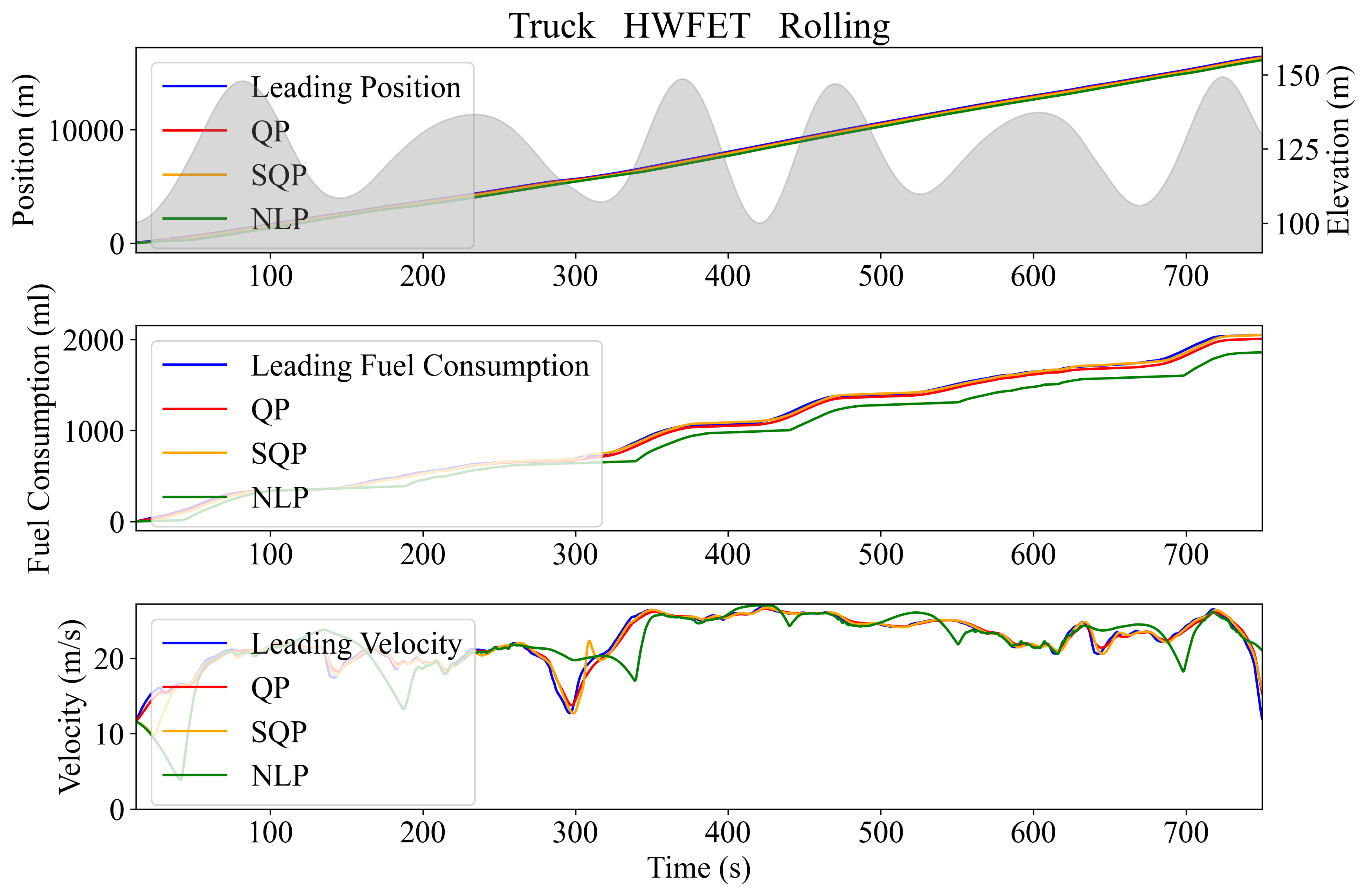}
        \caption{An experiment running on ``HWFET-Rolling''  profile.}
        \label{fig_rolling}
    \end{subfigure}
    
    \begin{subfigure}{.5\textwidth}
        \centering
        \includegraphics[width=\linewidth]{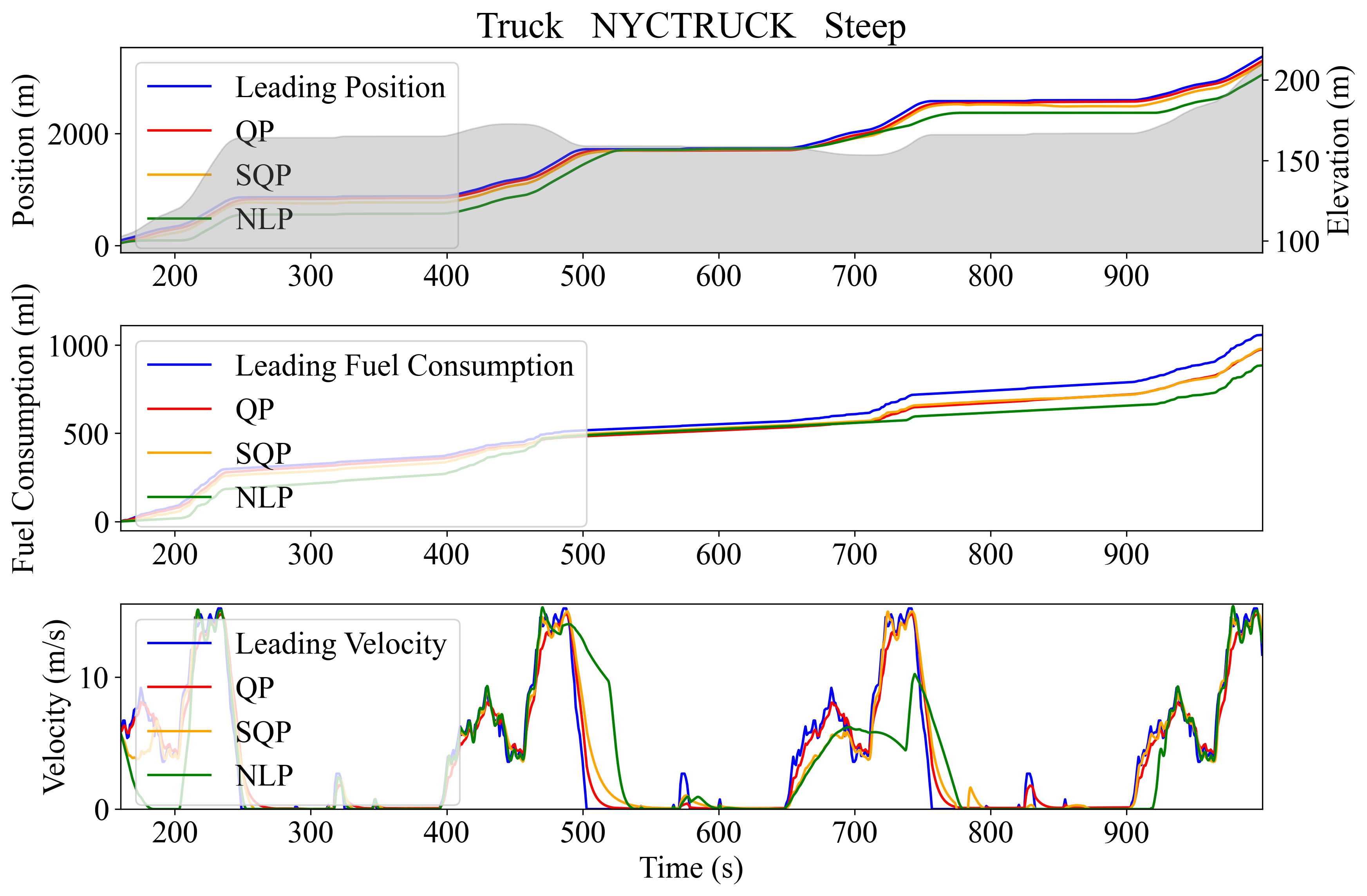}
        \caption{An experiment running on ``NYC\_TRUCK-Steep''  profile.}
        \label{fig_steep}
    \end{subfigure}

\caption{Example experiments, in the velocity graph, the leading agent in the blue line represents the standard driving cycle speed profile, and shaded fill-in denotes the elevation of the leading agent w.r.t. time coordinate. Fuel consumption of QP, SQP, and NLP agents are compared by lines in different colors, where NLP agents could achieve lower consumption.}
\label{fig_experiments}
\end{figure}

\subsection{Online Trajectory Optimization Framework}
We follow the previous studies~\cite{ocp_scp,acc_dp} to set an MPC-based online trajectory replanning framework that at each time step. Our proposed framework not only integrates a linear prediction of the leading car with $[s_l,v_l,a_{Vl}]$ at the current time step over a time horizon $H = N_T \cdot \Delta t $, but also predicts a $\theta(\cdot)$ trajectory over $H$. Adapting to AD online applications, slope profile is predicted w.r.t. time $t$  instead of $s$ used in the previous studies~\cite{ocp_scp, two_stage1}. Our proposed online trajectory framework is illustrated in Algorithm ~\ref{ag_framework}.

\begin{algorithm}
\caption{Online Trajectory Optimization}
\label{ag_framework}

Initialize  $\mathbf{G,X_l,X,U}$ $\leftarrow$ $\emptyset$ \\ 
Slover.$type$ $ \in \{NLP, SQP, QP\} $\\
\For{$t$ in $\{t_i \,|\, i \in [0,k], \,t_{i+1}-t_{i} = \Delta t \}$}{  
    $x_l^t, x^t \leftarrow UpdateStates(t)$ \\
    $\mathbf{X_l} \leftarrow PredictLeading(x_l)$~(\ref{eq_sv}), $\mathbf{X_l}[0] = x_l^t$ \;  % Fixed X_l(t) to X_l[t] if it's meant to be indexing
    \eIf{$\mathbf{X}= \emptyset$}{
        $d_{init} \leftarrow s_l^t - s^t$, \, s.t. \, $s_l^t \in x_l^t \quad s^t \in x^t $ \\
        $\mathbf{G} \leftarrow GetSlopeProfile(\mathbf{S_l}-d_{init}), \, \mathbf{S_l} \in \mathbf{X_l}$ \\
    }{
        $\mathbf{S}_G \leftarrow Concat(\mathbf{S}[1:N_T], \mathbf{S}[N_T])$ \\
        $\mathbf{G} \leftarrow GetSlopeProfile(\mathbf{S}_G )$  
    }
    Solver$.Update(X_l, G, x(t))$ \\
    $\mathbf{X} \leftarrow$ Solver$.Solve(),$ s.t.~(\ref{eq_kinematic})(\ref{eq_dynamic})(\ref{eq_acc}), $\mathbf{X}$ limits\\
    \eIf{Solver.$type$ = QP}{
    Execute $\mathbf{A}[0]$ \\
    }{
    Execute $\mathbf{U}[0]$ and  $\mathbf{B}[0]$ \\
    }
    Agent states transition.
}
\end{algorithm}
% such that $s$ coordinates starting from the subsequent timestep ($\mathbf{S}[1]$, timestep for the current loop aligns sequence number 1 in the previous $\mathbf{X}$) will be used as reference for slope profile in the new loop, to maintain the same length, each guess will duplicate the last element at the sequence number $N_T$. 

In line 5, we use the current leading agent states to linearly predict their trajectory by the kinematic equation~(\ref{eq_sv}). To predict the slope from lines 6-11, we first initialize a slope reference with $s_l$ and an initial intervehicle distance $d_{init}$, then update this reference by rolling out the solved ego trajectory $X$. With this $s$ reference, slope profile $\mathbf{G}$ can be precisely predicted w.r.t. time. Note that in line 13, since the state variables are overlapped with the control variables, we would return the whole state trajectory. In lines 15 and 17, different control commands for different settings are executed with its first element based on an MPC policy, then the program will update agent states and restart the loop.

\subsection{Experiment Setup}
To comprehensively test our proposed online trajectory optimization framework with three methods ``QP'', ``SQP'', and ``NLP'', we build up a testbed consisting of six driving cycles and three slope profiles. Two highway scenarios including ``HWFET'', ``INDIA HWY'', and four urban scenarios including ``INDIA URBAN'', ``NYCC'', ``NYC TRUCK'', ``MANHATTAN'' driving cycles are incorporated; By using Equation~(\ref{eq_theta}), ``flat'',``rolling'', and ``steep'' road profiles are set up with ($\theta_0 = 0, a_g = 0$), ($\theta_0 = 0, a_{gi} \in \{0.04,0.02\}, l_{wi} \in \{2870,2136\}$), and ($\theta_0 = 0.02, a_{gi} \in \{0.05,0.02, 0.01\}, l_{wi} \in \{2380,1860,1430\}$). An example of individual experiments running on a single driving cycle and road profile is shown in Fig.~\ref{fig_experiments}. Comprehensive tests are conducted with six-cycle, three-road, two-vehicle, and three-method settings. Besides the common compared metrics like travel distance, fuel consumption, and fuel efficiency, this study also evaluates the energy-saving policy on the following factors. \emph{(i) Planning and Prediction Horizon Length.} The planned trajectory for the ego agent and the prediction horizon of the traffic-slope profile share the same length,  5 and 10 seconds horizon (50 and 100 steps with $\Delta t$ set as $0.1$ second) are selected to compare which is reflected in the agent name in Table~\ref{tab_experiments}. Horizons shorter like 2 seconds have been tested to have inefficient results such that it is skipped due to page limits. \emph{(ii) Slope Prediction.} We also analyze the impact of usage of slope prediction or not, where only the current slope of the ego agent $\theta(s^{t_0})$ will be used for planning if not using a slope prediction $[\theta(s^{t_0}),\theta(s^{t_1}),...,\theta(s^{t_{N_T}})]$. This is reflected in Table~\ref{tab_experiments} where the agent name contains an ``F" to denote a false boolean for using slope prediction. In Table~\ref{tab_experiments}, original $6\times3\times2\times3\times2\text{(horizon)}\times2\text{(slope prediction)} = 432 $ tests are incorporated in a compact demonstration that results from various cycles and road profiles are summed and combined to do a total analysis due to page limits. QP, SQP, and NLP methods are implemented in Python with the OSQP~\cite{osqp} and IPOPT~\cite{ipopt}, within Casadi~\cite{casadi} symbolic expressions. Solving time tests are conducted on: A.~Windows PC i7 CPU and B.~Ubuntu Nvidia Xavier (embedded machine), and the corresponding solving time is also reflected in Table~\ref{tab_experiments}. Weights for each method are preoptimized by extensive tests and are listed in the Appendix.

\begin{table}[h]
\caption{NLP vs QP}
\centering
\begin{tabular}{cccc}
\hline
\makecell[c]{\textbf{Agent} \\ \textbf{ID}} & 
\makecell[c]{\textbf{Speed} \\ \textbf{Lose (\%)} } & 
\makecell[c]{\textbf{Fuel Efficiency} \\ \textbf{Improvement(\%)}} & 
\makecell[c]{\textbf{Solving Time} \\ \textbf{Multiples A}} \\ \hline
Sedan-QP-5 & 0.00 & 0.00 & 1.00 \\
Sedan-NLP-5 & 2.08 & 3.71 & 8.73 \\ \hline
Truck-QP-5 & 0.00 & 0.00 & 1.00 \\
Truck-NLP-5 & 2.49 & \color{darkgreen}{\textbf{7.15}} & 9.28 \\ \hline
\end{tabular}
\label{tab_nlp_qp}
\end{table}

\subsection{Results Analysis}
\subsubsection{Planning-Predction  Horizon and Usage of Slope Prediction} The results show that a medium horizon of 5 seconds could outperform 10 seconds and 2 seconds~(not shown) for both vehicles and all methods, where the model accuracy is satisfied. Using slope prediction would slightly improve fuel efficiency but not trivial for NLP (slope not included in QP), however, missing the slope prediction could cause SQP a higher inaccuracy leading to safety constraints violations, which is reflected in the gray-marked rows (some SQP tests failed due to constraint violations).
\subsubsection{Solving Time} 
All implemented methods can be solved efficiently on both devices, where SQP could run fast iterations with QP solver except for some fails, NLP takes 8-9 times of solving time of QP in Table~\ref{tab_nlp_qp}, while still maintaining an on-board (embedded device B) and online solving capability with a computation frequency of about 25 Hz. We believe this could satisfy an AD operation under normal working conditions. Device B requires approximately 3.76 times the computing time of Device A.
\subsubsection{Velocity-Energy Efficiency Tradeoff} Here we compare the proposed NLP method with the QP method (model-aware vs model-agnostic) on speed loss and fuel efficiency improvement in Table~\ref{tab_nlp_qp} shows that both vehicles could improve fuel efficiency without significantly reducing average speed. Specifically, the truck achieves a more substantial improvement of 7.15\% in energy savings. Given the total U.S. trucking industry's annual 327.48 billion trips~\cite{truck_mileage}, such an improvement could potentially generate \$6.14 billion in revenue.

\section{Conclusion}
\label{sec_conclusion}
% This study proposes an energy-model-aware direct NLP trajectory optimization method that bridges the gaps in the model-agnostic AD trajectory planning methods. This not only guarantees the safety in an ACC operation but also achieves 7.15\% higher energy efficiency on a tested truck model compared to a QP model-agnostic method widely used in AD studies without too much compromise on velocity. 

This study introduces a direct NLP trajectory optimization method within an energy-model-aware paradigm, effectively addressing the limitations observed in model-agnostic trajectory planning approaches. By incorporating a precise energy model directly into the trajectory planning, this method not only ensures enhanced safety during ACC operations but also demonstrates a significant increase in energy efficiency for both a tested sedan and truck. Specifically, the truck achieves a 7.15\% improvement in fuel saving~(\$6.14 billion equivalently) compared to the conventional QP model-agnostic method prevalent in AD research, without substantial loss of time efficiency, highlighting the method’s practical applicability. 

\appendix

\begin{table}[h]
\caption{Vehicle Parameters}
\centering % Center the table within the text block
\begin{tabular}{c|c|c||c|c|c} % Define a table with 5 columns
\hline
\textbf{Param} & \textbf{Sedan} & \textbf{Truck} & \textbf{Param} & \textbf{Sedan} & \textbf{Truck}\\
\hline
$M$   & 1200 & 4800 & $o_1$ & 1.0254e-2 & 9.0901e-3 \\
$A_v$  &  2.5  & 2.5 & $o_2$ & -9.2812e-4 & 3.7574e-8 \\
$\rho$   & 1.184 & 1.184 &$o_3$  & 2.154e-5 & 3.4935e-8 \\
$C_d$   & 0.32 & 0.6 & $o_4$  & -4.2427e-7 & 2.4230e-4 \\  
$\mu$  & 0.015  & 0.006 & $c_0$ &  0.07224 & 1.6550e-1           \\
$g$     & 9.81 & 9.81   &  $c_1$ & 0.09681 & 3.6070e-1              \\
$o_0$ &  1.4627e-1 &3.351e-1 & $c_2$ & 1.0750e-3 &  2.4223e-4       \\
\hline
\end{tabular}
\label{tab_vehicle_parameters}
\end{table}

\begin{table}[h]
\caption{Optimzation Parameters}
\centering % Center the table within the text block
\begin{tabular}{c|c|c|c||c|c|c} % Define a table with 5 columns
\hline
\textbf{Param} & \textbf{QP} & \textbf{SQP} & \textbf{NLP} & \textbf{Param} & \textbf{Sedan} & \textbf{Truck} \\
\hline
$w_1$   & 0.1 & 0.1 & 0.1 & $d_{init}$ & 50 & 50\\
$w_2$  &  2  & 5 & 5 & $v_{max}$ & 30 & 27 \\
$w_3$   & - & 2 & 10  & ${a_v}_{max}$ & 2.0 & 2.0 \\
$d_{min}$   & 10 & 10 & 10  & $b_{max}$ & 5.0 & 5.0 \\  
$d_{max}$  & 100  & 200 & 100  &  $ u_{max}$ & 9.0 & 3.0           \\

\hline
\end{tabular}
\label{tab_vehicle_parameters}
\end{table}

Sedan fuel model variables in~\cite{down_slope} are converted to $v, u$. View codes at this \href{https://github.com/Zhaofeng-Tian/Energy-Model-Aware-Trajectory-Optimization.}{link}.

\bibliographystyle{IEEEtran}
\bibliography{main}

\end{document}